# Directed Cyclic Graphical Representations of Feedback Models

Peter Spirtes
Department of Philosophy
Carnegie Mellon University
Pittsburgh, PA 15213

## Abstract

The use of directed acyclic graphs (DAGs) to represent conditional independence relations among random variables has proved fruitful in a variety of ways. Recursive structural equation models are one kind of DAG model. However, non-recursive structural equation models of the kinds used to model economic processes are naturally represented by directeed cyclic graphs (DCG). For linear systems associated with DCGs with independent errors, a characterisation of conditional independence constraints is obtained, and it is shown that the result generalizes in a natural way to systems in which the error variables or noises are statistically dependent. For non-linear systems with independent errors a sufficient condition for conditional independence of variables in associated distributions is obtained.

## 1. INTRODUCTION

The introduction of statistical models represented by directed acyclic graphs (DAGs) has proved fruitful in the construction of expert systems, in allowing efficient updating algorithms that take advantage of conditional independence relations (Pearl, 1988, Lauritzen et. al., 1993), and in inferring causal structure from conditional independence relations (Spirtes and Glymour, 1991, Spirtes, Glymour and Scheines, 1993, Pearl and Verma, 1991, Cooper, 1992). As a framework for representing the combination of causal and statistical hypotheses, DAG models have shed light on a number of issues in statistics ranging from Simpson's Paradox to experimental design (Spirtes, Glymour and Scheines, 1993). The relations of DAGs with statistical constraints, and the equivalence and distinguishability properties of DAG models, are now well understood, and their characterization and computation involves three properties connecting graphical structure and probability distributions: (i) a local directed Markov property, (ii) a global directed Markov property, (iii) and factorizations of joint densities according to the structure of a graph (Lauritizen et al., 1990).

Recursive structural equation models are one kind of DAG model. However, non-recursive structural equation models are not DAG models, and are instead naturally represented by directed *cyclic* graphs in which a finite series of edges representing influence leads from a vertex representing a variable back to that same vertex. Such graphs have been used to model feedback systems in electrical engineering (Mason, 1953, 1956), and to represent economic processes (Haavelmo, 1943, Goldberger, 1973). In contrast to the acyclic case, almost nothing general is known about how directed cyclic graphs (DCGs) represent conditional independence constraints, or about their equivalence or identifiability properties, or about characterizing classes of DCGs from conditional independence relations or other statistical constraints. This paper addresses the first of these problems, which is a prerequisite for the others. The issues turn on how the relations among properties (i), (ii) and (iii) essential to the acyclic case generalize—or more typically fail to generalize—to directed cyclic graphs and associated families of distributions. It will be shown that when DCGs are interpreted by analogy with DAGs as representing functional dependencies with independently distributed noises or "error terms," the equivalence of the fundamental global and local Markov conditions characteristc of DAGs no longer holds, even in linear systems, and in non-linear systems both Markov properties may fail. For linear systems associated with DCGs with independent errors or noises, a characterisation of conditional independence constraints is obtained, and it is shown that the result generalizes in a natural way to systems in which the error variables or noises are statistically dependent. For non-linear systems with independent errors a sufficient condition for conditional independence of variables in associated distributions is obtained.

The remainder of this paper is organized as follows: Section 2 defines relevant mathematical ideas and gives some necessary technical results on DAGs and DCGs. Section 3 obtains results for non-recursive linear structural equations models. Section 4 treats non-linear models of the same kind.

## 2. DIRECTED GRAPHS

I place sets of variables and defined terms in boldface, and individual variables in italics. A **directed graph** is an ordered pair of a finite set of vertices **V**, and a set of directed edges **E**. A directed edge from $A$ to $B$ is an ordered pair of distinct vertices $<A,B>$ in **V** in which $A$ is the **tail** of the edge and $B$ is the **head**; the edge is **out of** $A$ and **into** $B$, and $A$ is **parent** of $B$ and $B$ is a **child** of $A$. A sequence of distinct edges $<E_1,...,E_n>$ in $G$ is an **undirected path** if and only if there exists a sequence of vertices $<V_1,..., V_{n+1}>$ such that for $1 \le i \le n$ either $<V_i, V_{i+1}> = E_i$ or $<V_{i+1},V_i> = E_i$. A path $U$ is **acyclic** if no vertex occurring on an edge in the path occurs more than once. A sequence of distinct edges $<E_1,..., E_n>$ in $G$ is a **directed path** if and only if there exists a sequence of vertices $<V_1,..., V_{n+1}>$ such that for $1 \le i \le n$, $<V_i, V_{i+1}> = E_i$. If there is an acyclic directed path from $A$ to $B$ or $B = A$ then $A$ is an **ancestor** of $B$, and $B$ is a **descendant** of $A$. A directed graph is **acyclic** if and only if it contains no directed cyclic paths.[1]

A **directed acyclic graph** (DAG) $G$ with a set of vertices **V** can be given two distinct interpretations. On the one hand, such graphs can be used to represent causal relations between variables, where an edge from $A$ to $B$ in $G$ means that $A$ is a direct cause of $B$ relative to **V**. A **causal** graph is a DAG given such an interpretation. On the other hand, a DAG with a set of vertices **V** can also represent a set of probability measures over **V**. Following the terminology of Lauritzen et. al. (1990) say that a probability measure over a set of variables **V** satisfies the **local directed Markov property** for a DAG $G$ with vertices **V** if and only if for every $W$ in **V**, $W$ is independent of **V**\(**Descendants**$(W,G)$ ∪ **Parents**$(W,G)$) given **Parents**$(W,G)$, where **Parents**$(W,G)$ is the set of parents of $W$ in $G$, and **Descendants**$(W,G)$ is the set of descendants of $W$ in $G$. A DAG $G$ **represents** the set of probability measures which satisfy the local directed Markov property for $G$. The use of DAGs to simultaneously represent a set of causal hypotheses and a family of probability distributions extends back to the path diagrams introduced by Sewell Wright (1934). Variants of DAG models were introduced in the 1980's in Wermuth (1980), Wermuth and Lauritzen (1983), Kiiveri, Speed, and Carlin (1984), Kiiveri and Speed (1982), and Pearl (1988).[2]

Lauritzen et. al. also define a **global directed Markov property** that is equivalent to the local directed Markov property for DAGs. Several preliminary notions are required. Let **An**(**X**,$G$) be the set of ancestors of members of **X** in $G$. Let $G($**X**$)$ be the subgraph of $G$ that contains only vertices in **X**, with an edge from $A$ to $B$ in **X** if and only if there is an edge from $A$ to $B$ in $G$. $G^M$ **moralizes** a directed graph $G$ if and only if $G^M$ is an undirected graph with the same vertices as $G$, and a pair of vertices $X$ and $Y$ are adjacent in $G^M$ if and only if either $X$ and $Y$ are adjacent in $G$, or they have a common child in $G$. In an undirected graph $G$, if **X**, **Y**, and **Z** are disjoint then **X** is **separated** from **Y** given **Z** if and only if every undirected path between a member of **X** and a member of **Y** contains a member of **Z**. If **X**, **Y** and **Z** are disjoint sets of variables, **X** and **Y** are **d-separated** given **Z** in a directed graph $G$ just when **X** and **Y** are separated given **Z** in $G^M($**An**$($**X** ∪ **Y** ∪ **Z**$,G))$. The relation defined here was defined in Lauritzen et al. (1990). "d-separation" is a graphical relation introduced by Pearl (1986). Since Lauritzen et. al. (1990) proved that their graphical relation is equivalent to Pearl's for acyclic graphs, and the proof is readily extended to the cyclic case, I will also use "d-separation" to refer to the graphical relation just described. Now the definition: A probability measure over **V** satisfies the **global directed Markov property** for DAG $G$ if and only if for any three disjoint sets of variables **X**, **Y**, and **Z** included in **V**, if **X** is d-separated from **Y** given **Z**, then **X** is independent of **Y** given **Z**. Lauritzen et. al. (1990) shows that the global and local directed Markov properties are equivalent in DAGs, even when the probability distributions represented have no density function. In section 2, I show that the local and global directed Markov properties are not equivalent for cyclic directed graphs.

The following lemmas relate the global directed Markov property to factorizations of a density function. Denote a density function over **V** by $f($**V**$)$, where for any subset **X** of **V**, $f($**X**$)$ denotes the marginal of $f($**V**$)$. If $f($**V**$)$ is the density function for a probability measure $P$ over a set of variables **V**, say that $P$ **factors according to directed graph** $G$ with vertices **V** if and only if for every subset **X** of **V**,

$$P = f \bullet \mu$$
$$f(\mathbf{An}(\mathbf{X},G)) = \prod_{V \in \mathbf{An}(\mathbf{X},G)} g_V(V, \mathbf{Parents}(V,G))$$

where $g_V$ is a non-negative function, and $\mu$ is a product measure. The following result was proved in Lauritzen et. al. (1990).

**Lemma 1:** If **V** is a set of random variables with a probability measure $P$ that is absolutely continuous with respect to a product measure $\mu$, then $P$ factors according to DAG $G$ if and only if $P$ satisfies the global directed Markov property for $G$.

As in the case of acyclic graphs, the existence of a factorization according to a cyclic directed graph $G$ does entail that a measure satisfies the global directed Markov property for $G$. The proof given in Lauritzen et. al. (1990) for the acyclic case carries over essentially unchanged for the cyclic case.

---

[1] An undirected path is often defined as a sequence of vertices rather than a sequence of edges. The two definitions are essentially equivalent for acyclic directed graphs, because a pair of vertices can be identified with a unique edge in the graph. However, a cyclic graph may contain more than one edge between a pair of vertices. In that case it is no longer possible to identify a pair of vertices with a unique edge.

[2] It is often the case that some further restrictions are placed on the set of distributions represented by a DAG. For example, one could also require the Minimality Condition, i.e. that for any distribution $P$ represented by $G$, $P$ does not satisfy the local directed Markov Condition for any proper subgraph of $G$. This condition, and others are discussed in Pearl(1988) and Spirtes, Glymour, and Scheines(1993). We will not consider such further restrictions here.

**Lemma 2:** If **V** is a set of random variables with a probability measure $P$ that factors according to directed (cyclic or acyclic) graph $G$, then $P$ satisfies the global directed Markov property for $G$.

However, unlike the case of acyclic graphs, if a probability measure over a set of variable **V** satisfies the global directed Markov property for cyclic graph $G$ and has a density function $f(\mathbf{V})$, it does not follow that $f(\mathbf{V})$ factors according to $G$.

The following weaker result relating factorization of densities and the global directed Markov property does hold for both cyclic and acyclic directed graphs.

**Lemma 3:** If **V** is a set of random variables with a probability measure $P$ that that is absolutely continuous with respect to a product measure $\mu$, has a positive density function $f(\mathbf{V})$, and satisfies the global directed Markov property for directed (cyclic or acyclic) graph $G$, then $f(\mathbf{V})$ factors according to $G$.

## 3. NON-RECURSIVE LINEAR STRUCTURAL EQUATION MODELS

The problem considered in this section is to investigate the generalization of the Markov properties to linear, non-recursive structural equation models. First we must relate the social scientific terminology to graphical representations, and clarify the questions.

Linear structural equation models (which, following the terminology of Bollen (1989), will be referred to as linear SEMs) can also be represented as directed graph models. In a linear SEM the random variables are divided into two disjoint sets, the error terms and the non-error terms. Corresponding to each non-error random variable $V$ is a unique error term $\varepsilon_V$. A linear SEM contains a set of linear equations in which each non-error random variable $V$ is written as a linear function of other non-error random variables and $\varepsilon_V$. A linear SEM also specifies a joint distribution over the error terms. So, for example, the following is a linear SEM, where $a$ and $b$ are real constant linear coefficients, $\varepsilon_X, \varepsilon_Y$, and $\varepsilon_Z$ are jointly independent "error terms", and $X, Y, Z$, are random variables:

$$X = a\,Y + \varepsilon_X$$
$$Y = b\,Z + \varepsilon_Y$$
$$Z = \varepsilon_Z$$

$\varepsilon_X, \varepsilon_Y, \varepsilon_Z$ are jointly independent and normally distributed

The directed graph of a linear SEM with uncorrelated errors is written with the convention that an edge does not appear if and only if the corresponding entry in the coefficient matrix is zero; the graph does not contain the error terms. Figure 1 is the DAG that represents the SEM shown above. A linear SEM is **recursive** if and only if its directed graph is acyclic.

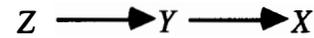

Figure 1: Example of Recursive SEM

Initially I will consider only linear SEMs in which the error terms are jointly independent, but we will see that in the linear case in an important sense nothing is lost by this restriction: a linear SEM with dependent errors generates the same restrictions on the covariance matrix as does some linear SEM with extra variables and independent errors. Further, such an SEM with extra variables can always be found such that the subgraph of that SEM on the original variables is the same as the original graph.

A linear SEM containing disjoint sets of variables **X**, **Y**, and **Z linearly entails** that **X** is independent of **Y** given **Z** if and only if **X** is independent of **Y** given **Z** for all values of the non-zero linear coefficients and all distributions of the exogenous variables in which they are jointly independent and have positive variances. Let $\rho_{XY\cdot Z}$ be the partial correlation of $X$ and $Y$ given **Z**. A linear SEM containing $X$, $Y$, and **Z**, where $X \neq Y$ and $X$ and $Y$ are not in **Z**, **linearly entails** that $\rho_{XY\cdot Z} = 0$ if and only $\rho_{XY\cdot Z} = 0$ for all values of the non-zero linear coefficients and all distributions of the exogenous variables in which each pair of exogenous variables has zero correlation, each exogenous variable has positive variance, and in which $\rho_{XY\cdot Z}$ is defined. It follows from Kiiveri and Speed (1982) that if the error terms are jointly independent, then any distribution that forms a linear, recursive SEM with a directed graph $G$ satisfies the local directed Markov property for $G$. One can therefore apply d-separation to the DAG in a linear, recursive SEM to compute the conditional independencies and zero partial correlations it linearly entails. The d-separation relation provides a polynomial (in the number of vertices) time algorithm for deciding whether a given vanishing partial correlation is linearly entailed by a DAG.

Linear non-recursive structural equation models (linear SEMs) are commonly used in the econometrics literature to represent feedback processes that have reached equilibrium.[3] Corresponding to a set of non-recursive linear equations is a cyclic graph, as the following example from Whittaker (1990) illustrates.

$$X_1 = \varepsilon_{X1}$$
$$X_2 = \varepsilon_{X2}$$
$$X_3 = \beta_{31}X_1 + \beta_{34}X_4 + \varepsilon_{X3}$$
$$X_4 = \beta_{42}X_2 + \beta_{43}X_3 + \varepsilon_{X4}$$

$\varepsilon_{X1}, \varepsilon_{X2}, \varepsilon_{X3}, \varepsilon_{X4}$ are jointly independent and normally distributed

---

[3] Cox and Wermuth (1993), Wermuth and Lauritzen(1990) and (indirectly) Frydenberg(1990) consider a class of non-recursive linear models they call *block recursive*. The block recursive models overlap the class of SEMs, but they are neither properly included in that class, nor properly include it. Frydenberg (1990) presents necessary and sufficient conditions for the equivalence of two block recursive models.

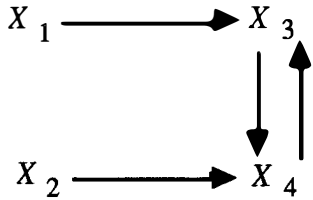

Figure 2: Example of Non-recursive SEM

In DAGs the global directed Markov property entails the local directed Markov property, because a variable $V$ is d-separated from its non-parental non-descendants given its parents. This is not always the case in cyclic graphs. For example, in figure 4, $X_4$ is not d-separated from its non-parental non-descendant $X_1$ given its parents $X_2$ and $X_3$, so the local directed Markov property does not hold.

(Note that this use of cyclic directed graphs to represent feedback processes represents an extension of the causal interpretation of directed graphs. The causal structure corresponding to Figure 2 is described by an infinite acyclic directed graph containing each variable indexed by time. The cyclic graph can be viewed as a compact representation of such a causal graph. I am indebted to C. Glymour for pointing out that the local Markov condition fails in Whittaker's model. Indeed, there is no acyclic graph (even with additional variables) that linearly entails all and conditional independence relations linearly entailed by Figure 2, although Thomas Richardson has pointed out that the directed cyclic graph of Figure 2 is equivalent to one in which the edges from $X_1$ to $X_3$ and $X_2$ to $X_4$ are replaced, respectively, by edges from $X_1$ to $X_4$ and from $X_2$ to $X_3$.)

**Theorem 1:** The probability measure $P$ of a linear SEM $L$ (recursive or non-recursive) with jointly independent error terms satisfies the global directed Markov property for the directed (cyclic or acyclic) graph $G$ of $L$, i.e. if $\mathbf{X}$, $\mathbf{Y}$, and $\mathbf{Z}$ are disjoint sets of variables in $G$ and $\mathbf{X}$ is d-separated from $\mathbf{Y}$ given $\mathbf{Z}$ in $G$, then $\mathbf{X}$ and $\mathbf{Y}$ are independent given $\mathbf{Z}$ in $P$.[4]

**Theorem 2:** In a linear SEM $L$ with jointly independent error terms and directed (cyclic or acyclic) graph $G$ containing disjoint sets of variables $\mathbf{X}$, $\mathbf{Y}$ and $\mathbf{Z}$, if $\mathbf{X}$ is not d-separated from $\mathbf{Y}$ given $\mathbf{Z}$ then $L$ does not linearly entail that $\mathbf{X}$ is independent of $\mathbf{Y}$ given $\mathbf{Z}$.

Applying Theorems 1 and 2 to the directed graph in Figure 2, only two conditional independence relations (and their consequences) are entailed: $X_1$ is independent of $X_2$, and $X_1$ is independent of $X_2$ given $X_3$ and $X_4$.

**Theorem 3:** In a linear SEM $L$ with jointly independent error terms and (cyclic or acyclic) directed graph $G$ containing $X$, $Y$ and $\mathbf{Z}$, where $X \neq Y$ and $\mathbf{Z}$ does not contain $X$ or $Y$, $X$ is d-separated from $Y$ given $\mathbf{Z}$ if and only if $L$ linearly entails that $\rho_{XY \cdot \mathbf{Z}} = 0$.

As in the acyclic case, d-separation provides a polynomial time procedure for deciding whether cyclic graphs entail a conditonal independence or vanishing partial correlation

Theorem 3 can be used to relax the restriction that the error terms in an a linear SEM $L$ be jointly independent. We will slightly extend the definition of linear entailment. Assume that some pairs of exogenous variables may have correlations fixed at zero. A linear SEM containing $X$, $Y$, and $\mathbf{Z}$, where $X \neq Y$ and $X$ and $Y$ are not in $\mathbf{Z}$, **linearly entails** that $\rho_{XY \cdot \mathbf{Z}} = 0$ if and only $\rho_{XY \cdot \mathbf{Z}} = 0$ for all values of the non-zero linear coefficients and for each distribution over the exogenous variables in which pairs of exogenous variables fixed at zero correlation keep a zero correlation, each exogenous variable has positive variance, and in which $\rho_{XY \cdot \mathbf{Z}}$ is defined. If $\varepsilon_X$ and $\varepsilon_Y$ are not independent in linear SEM $L$ with a non-singular covariance matrix, there is a linear SEM $L'$ with independent error terms such that the marginal distribution of $L'$ over the variables in $L$ has the same covariance matrix as $L$. Form the graph $G'$ of $L'$ from the graph $G$ of $L$ in the following way. Add a latent variable $T$ to $G$, and add edges from $T$ to $X$ and $Y$. In $L'$, modify the equation for $X$ by making it a linear functions of the parents of $X$ (including $T$) in $G'$, and replace $\varepsilon_X$ by $\varepsilon'_X$; modify the equation for $Y$ in an analogous way. There always exist linear coefficients and distributions over $T$ and the new error terms such that the marginal covariance matrix for $L'$ is equal to the covariance matrix of $L$, and $\varepsilon'_X$ and $\varepsilon'_Y$ are independent. The process can be repeated for each pair of variables with correlated errors in $L$. Hence the zero partial correlations entailed by $L$ can be derived by applying Theorem 3 to the graph of $L'$. Figure 3 illustrates this process. The set of variables $\mathbf{V}$ in the graph on the left is $\{X_1, X_2, X_3, X_4\}$. The graph on the left correlates the errors between $X_1$ and $X_2$ (indicated by the undirected edges between them.) The graph on the right has no correlated errors, but does have a latent variable $T$ that is a parent of $X_1$ and $X_2$. The two graphs linearly entail the same zero partial correlations involving only variables in $\mathbf{V}$ (in this case they both entail no non-trivial zero partial correlations).

$X_3 = a \times X_2 + b \times X_4 + \varepsilon_3$   $X_3 = a \times X_2 + b \times X_4 + \varepsilon_3$
$X_4 = c \times X_1 + d \times X_3 + \varepsilon_4$   $X_4 = c \times X_1 + d \times X_3 + \varepsilon_4$
$X_1 = \varepsilon_1$                              $X_1 = e \times T + \varepsilon'_1$
$X_2 = \varepsilon_2$                              $X_2 = f \times T + \varepsilon'_2$
$\varepsilon_1$ and $\varepsilon_2$ correlated    $\varepsilon'_1$ and $\varepsilon'_2$ uncorrelated

## 4. NON-LINEAR STRUCTURAL EQUATION MODELS

A linear SEM is a special case of a more general kind of SEM in which the equations relating a given variable to other variables and a unique error term need not be linear. In a SEM the random variables are divided into two

---

[4] This theorem has been independently proved by Jan Koster of the Erasmus University Rotterdam, in a paper which has not yet been published but has been submitted to Statistical Science.

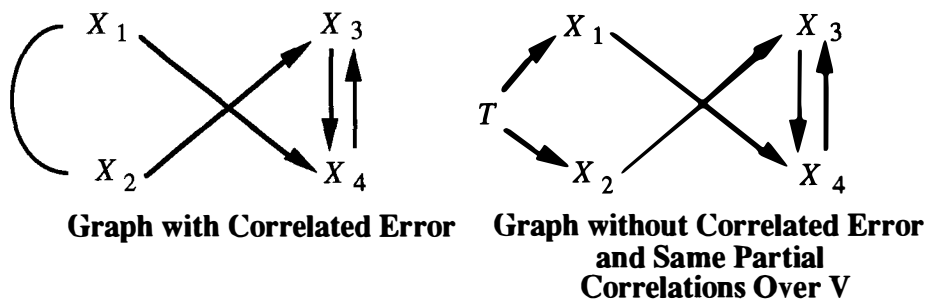

**Graph with Correlated Error**    **Graph without Correlated Error and Same Partial Correlations Over V**

Figure 3: Correlated Errors

disjoint sets, the error terms and the non-error terms. Corresponding to each non-error random variable $V$ is a unique error term $\varepsilon_V$. A SEM contains a set of equations in which each non-error random variable $V$ is written as a measureable function of other non-error random variables and $\varepsilon_V$. The convention is that in the directed graph of a SEM there is an edge from $A$ to $B$ if and only if $B$ is an argument in the function for $A$. As in the linear case, I will still assume that density functions exist for both the probabilty measure over the error terms and the non-error terms, each non-error term $V$ is a function of the error terms of its ancestors in $G$, each $\varepsilon_V$ is a function of $V$ and its parents in $G$ (which will be the case if the errors are additive or multiplicative), the Jacobian of the transformation between the error terms and the non-error terms is well-defined, and the error terms are jointly independent. Call such a set of equations and its associated graph a **pseudo-indeterministic** SEM. A directed graph $G$ **pseudo-indeterministically entails** that $\mathbf{X}$ is independent of $\mathbf{Y}$ given $\mathbf{Z}$ if and only if in every pseudo-indeterministic SEM with graph $G$, $\mathbf{X}$ is independent of $\mathbf{Y}$ given $\mathbf{Z}$.

This section establishes that d-separation is a necessary condition for a DAG to pseudo-indeterministically entail a conditional independence relation, but in a cyclic directed graph d-separation may not be a sufficient condition for a DAG to pseudo-indeterministically entail a conditional independence relation. Instead, a different condition, yielding a polynomial time algorithm, is found to suffice for a cyclic direected graph to pseudo-indeterministically entail a conditional independence relation.

By Theorem 2, d-separation is a necessary condition for a conditional independence claim to be entailed by an SEM. The following remarks show d-separation is also sufficient for acyclic SEMs, but not for cyclic SEMS.

**Theorem 4:** If $G$ is a DAG containing disjoint sets of variables $\mathbf{X}$, $\mathbf{Y}$ and $\mathbf{Z}$, $\mathbf{X}$ is d-separated from $\mathbf{Y}$ given $\mathbf{Z}$ if and only $L$ pseudo-indeterministically entails that $\mathbf{X}$ is independent of $\mathbf{Y}$ given $\mathbf{Z}$.

The following example gives a concrete illustration that there is a cyclic graph $G$ in which $X$ is d-separated from $Y$ given $\{Z,W\}$, but $G$ does not pseudo-indeterminstically entail that $X$ is independent of $Y$ given $\{Z,W\}$.

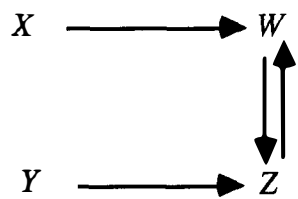

Figure 4: Graph G

$$X = \varepsilon_X$$
$$Y = \varepsilon_Y$$
$$Z = W \times Y + \varepsilon_Z$$
$$W = Z \times X + \varepsilon_W$$

$\varepsilon_X$, $\varepsilon_Y$, $\varepsilon_Z$, $\varepsilon_W$ with independent standard normal distributions

The transformation from $\varepsilon_X$, $\varepsilon_Y$, $\varepsilon_Z$, $\varepsilon_W$ to $X, Y, Z, W$ is 1-1 except where $\varepsilon_X \times \varepsilon_Y = 1$ because

$$X = \varepsilon_X$$
$$Y = \varepsilon_Y$$
$$Z = \frac{\varepsilon_W \times \varepsilon_Y + \varepsilon_Z}{1-(\varepsilon_X \times \varepsilon_Y)}$$
$$W = \frac{\varepsilon_Z \times \varepsilon_X + \varepsilon_W}{1-(\varepsilon_X \times \varepsilon_Y)}$$

The Jacobean of the transformation from the $\varepsilon$'s is $1/(1 - X \times Y)$. Hence, transforming the joint normal density of the $\varepsilon$'s yields

$$f(X,Y,Z,W) =$$
$$\frac{1}{4\pi^2} \times \exp(\frac{-x^2}{2}) \times \exp(\frac{-y^2}{2}) \times$$
$$\exp(\frac{-(z - w \times y)^2}{2}) \times \exp(\frac{-(w - z \times x)^2}{2}) \times$$
$$\left|\frac{1}{1-(X \times Y)}\right|$$

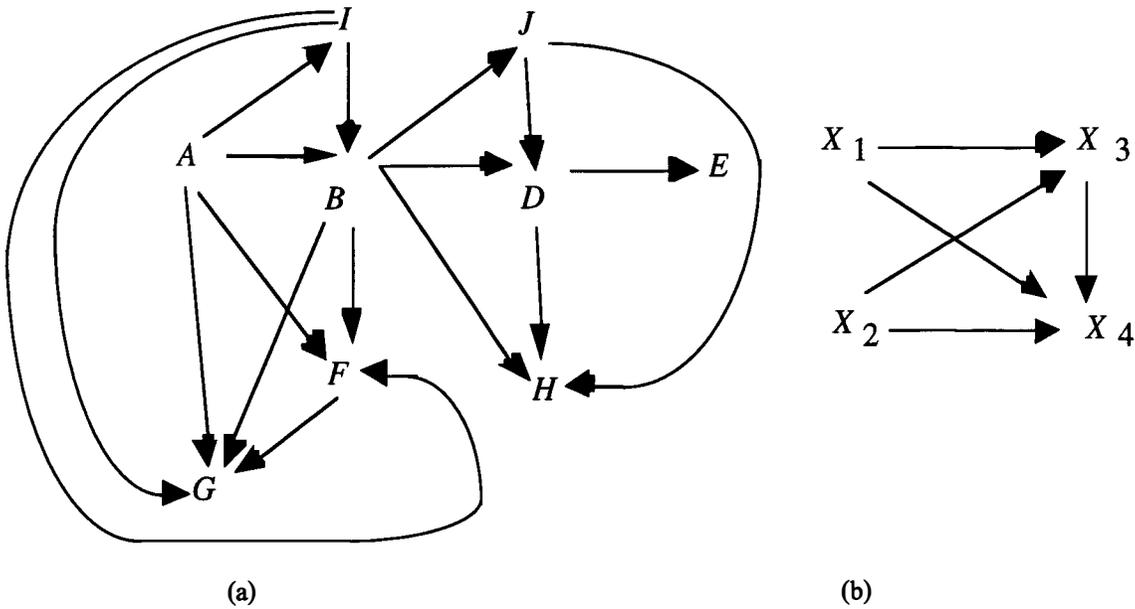

(a)                                                          (b)

Figure 6: Collapsed Graph

$X$ is not independent of $Y$ given $\{Z,W\}$ in this distribution because it is not possible to factor it into a product of terms, no one of which contains both $X$ and $Y$.

However, it is possible to modify the graphical representation of the functional relations in such a way that d-separation applied to the new graph does entail conditional independence. In a directed graph $G$, a **cycle** is a cyclic directed path, in which each vertex occurs on exactly two edges. A set of cycles $\mathbf{C}$ is a **cyclegroup** if and only if it is a smallest set of cycles such that for each cycle $C_1$ in $\mathbf{C}$, $\mathbf{C}$ contains the transitive closure of all of the cycles intersecting $C_1$, i.e. it contains all of the cycles that intersect $C_1$, all of the cycles that intersect cycles that intersect $C_1$, etc. For example, in figure 5, there are two distinct cyclegroups: the first is $\{C_1, C_2, C_3\}$, and the second is $\{C_4, C_5\}$.

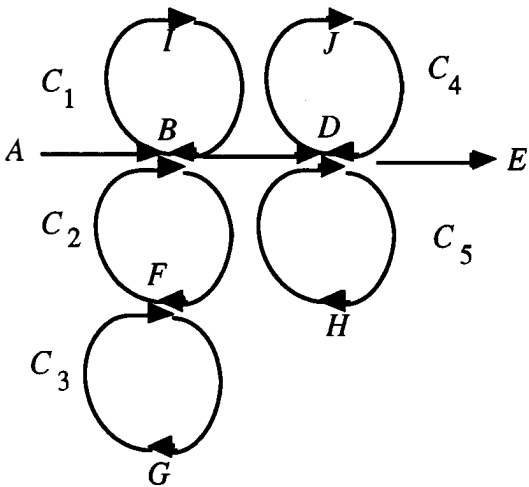

Figure 5: Cyclegroups

Let the set of all cycles in $G$ be **Cycles**($G$). If a vertex $V$ or an edge $<V,W>$ occurs in some set $\mathbf{C}$ of cycles, for brevity write $V \in \mathbf{C}$ or $<V,W> \in \mathbf{C}$ respectively, although strictly speaking neither a vertex nor an edge is a member of a set of cycles. Form the **collapsed graph** $G'$ from $G$ by the following operations on each cyclegroup:

1. remove all of the edges between members of the cyclegroup;

2. arbitrarily number the vertices in the cyclegroup;

3. add an edge from each lower number vertex to each higher number vertex;

4. for each parent $A$ of a member of the cyclegroup that is not itself in the cyclegroup, add an edge from $A$ to each member of the cyclegroup.

The procedure does not define a unique collapsed graph due to the arbitrariness of the numbering, but since all of the collapsed graphs share the same d-separation relations, it does not matter. Note that even if $G$ is a cyclic graph, the collapsed graph is acyclic. The collapsed graph can be generated in polynomial time.

**Theorem 5:** In an SEM with directed graph $G$ (cyclic or acyclic) and collapsed graph $G'$ containing disjoint sets of variables **X**, **Y** and **Z**, if **X** is d-separated from **Y** given **Z** in $G'$ then the SEM entails that **X** is independent of **Y** given **Z**.

A collapsed graph for the graph in figure 5 is shown in figure 6a, and a collapsed graph for the graph in figure 4 is shown in figure 6b.

I make the following conjecture:

**Conjecture**: Let $G$ (cyclic or acyclic) have collapsed graph $G'$ containing disjoint sets of variables **X**, **Y** and **Z**. If $G$ pseudo-indeterministically entails that **X** is independent of **Y** given **Z**, then in $G'$ **X** is d-separated from **Y** given **Z**.

## 5. CONCLUSION

These are a number of issues related to these results which may be of practical importance. Under what conditions, for example, are there results about conditional independence comparable to the equivalence of vanishing partial correlations in models with dependent errors and latent variable models with independent errors? There are polynomial algorithms (Verma and Pearl, 1990, Frydenberg, 1990) for determining when two arbitrary directed acyclic graphs entail the same set of conditional independence relations. Richardson (1994, 1995) has shown that there is a polynomial time algorithm for determining when two arbitrary directed graphs (cyclic or acyclic) linearly entail the same set of conditional independence relations. There are polynomial time algorithms (Spirtes and Verma, 1992) for determining when two arbitrary directed acyclic graphs entail the same set of conditional independence relations over a common subset of variable **O**. Is there a polynomial time algorithm for determining when two arbitrary directed graphs (cyclic or acyclic) linearly entail the same set of conditional independence relations over a common subset of variables **O**? Assuming Markov properties hold and completely characterize the conditional independence facts in distributions considered, there are correct polynomial time algorithms for inferring features of (sparse) directed acyclic graphs from a probability distribution when there are no latent common causes (see Spirtes and Glymour, 1991, Cooper and Herskovitz, 1992). Richardson's characterization of equivalence for cyclic graphs suggests that there is an algorithm for discovering cyclic graphs, and that this algorithm will run in polynomial time on sparse graphs. There are similarly correct, but not polynomial, algorithms for inferring features of directed acyclic graphs from a probability distribution even when there may be latent common causes (see Spirtes, 1992 and Spirtes, Glymour and Scheines, 1993). Are there comparable algorithms for inferring features of directed graphs (cyclic or acyclic) from a probability distribution even when there may be latent common causes?

Finally, although non-recursive linear structural equation models are commonly used to represent the static equilibrium of a feedback process, further research is needed to determine what types of causal processes would arrive at this type of equilibrium state.

## Acknowledgements

Research for this paper was supported by the National Science Foundation through grant 9102169 and the Navy Personnel Research and Development Center and the Office of Naval Research through contract number N00014-93-1-0568. I am indebted to Clark Glymour, Richard Scheines, Christopher Meek, Thomas Richardson, and Marek Druzdzel for helpful conversations.